%% file: bare_jrnl_rv2.tex
\documentclass[journal]{IEEEtran}
\pdfoutput=1
\usepackage{graphicx}
\usepackage{amsmath}
\usepackage{amssymb}
\usepackage{enumitem}
\usepackage{xfrac}
\usepackage{commath}
\usepackage{multirow}
\usepackage{enumitem}
\usepackage{booktabs}
\usepackage{xcolor}
\usepackage{array}
\usepackage{tabularx}
\usepackage{algorithm}
\usepackage{algpseudocode}

\DeclareMathOperator*{\argmax}{arg\,max}

\newcommand{\update}[1]{{\textcolor{black}{#1}}}

\newcommand{\aqe}[1]{{\textcolor{black}{#1}}}

\ifCLASSINFOpdf
\else
\fi
\begin{document}
%
\title{Text-based Localization of Moments in a \\ Video Corpus}
%
%
%

\author{Sudipta Paul,~\IEEEmembership{Member,~IEEE}, Niluthpol Chowdhury Mithun,~\IEEEmembership{Member,~IEEE}, \\ and Amit K. Roy-Chowdhury,~\IEEEmembership{Fellow,~IEEE}
\thanks{$\bullet$ Sudipta Paul, and Amit~K.~Roy-Chowdhury are with the Department of Electrical and Computer Engineering, University of California, Riverside, CA, USA. Niluthpol Chowdhury Mithun is with SRI International, Princeton, NJ, USA. \  E-mails: (spaul007@ucr.edu, niluthpol.mithun@sri.com, amitrc@ece.ucr.edu)}}

%
%

\markboth{IEEE Transactions on Image Processing,~Vol.xx, No.xx, xxx~2020}%
{Sudipta \MakeLowercase{\textit{et al.}}: Bare Demo of IEEEtran.cls for IEEE Journals}
%



\maketitle

\begin{abstract}
\input{sections/N01_rv2_abstract}
\end{abstract}


%
\IEEEpeerreviewmaketitle

\input{sections/N02_rv2_introduction}

\input{sections/N03_rv2_related_work}

\input{sections/N04_rv2_methodology}

\input{sections/N05_rv2_experiment}

\input{sections/N06_rv2_conclusion}

\section*{Acknowledgment}
The work was partially supported by NSF grant IIS-1901379 and ONR grant N00014-19-1-2264.


\ifCLASSOPTIONcaptionsoff
  \newpage
\fi

\input{bare_jrnl_rv2.bbl}
\begin{IEEEbiography}[{\includegraphics[width=1in,height=1.4in,clip,keepaspectratio]{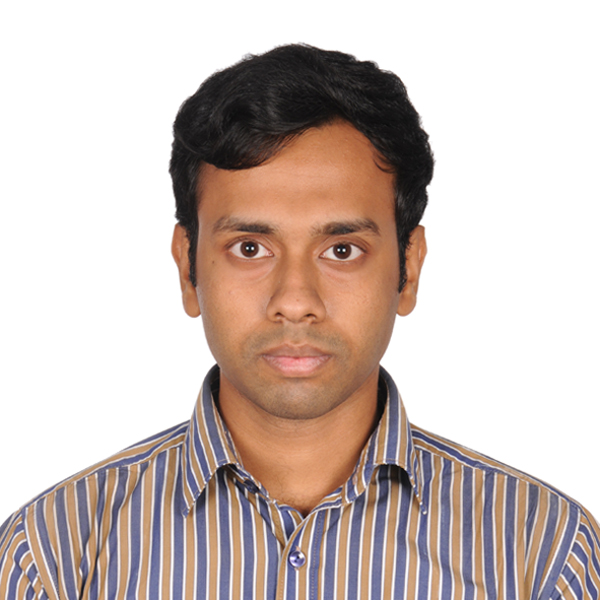}}]{Sudipta Paul}
received his Bachelor's degree in Electrical and Electronic Engineering from Bangladesh University of Engineering and Technology, Dhaka in 2016. He is currently pursuing his Ph.D. degree in the department of Electrical and Computer Engineering at University of California, Riverside. His main research interests include computer vision, machine learning, vision and language, and robust learning.
\end{IEEEbiography}

\begin{IEEEbiography}[{\includegraphics[width=1.15in,height=1.4in,clip,keepaspectratio]{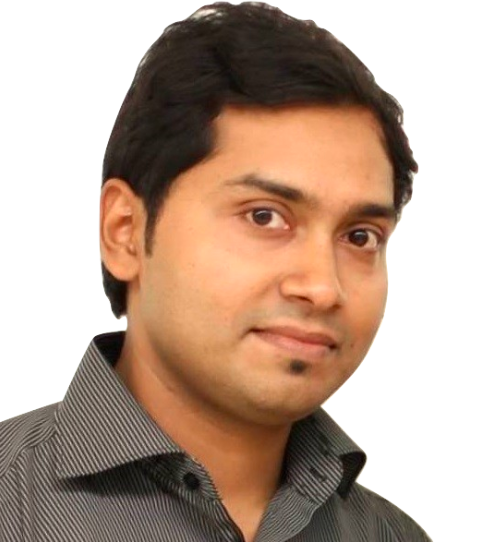}}]{Niluthpol Chowdhury Mithun} graduated from the University of California, Riverside with a Ph.D. in Electrical and Computer Engineering in 2019. Previously, he received his Bachelors and Masters degree from Bangladesh University of Engineering and Technology. He is currently an Advanced Computer Scientist at the Center for Vision Technologies, SRI International, Princeton. His broad research interest includes computer vision and machine learning with more focus on multimodal data analysis, weakly supervised learning and vision-based localization.
\end{IEEEbiography}
\vspace{-12cm}

\begin{IEEEbiography}[{\includegraphics[width=1.1in,height=1.1in,clip]{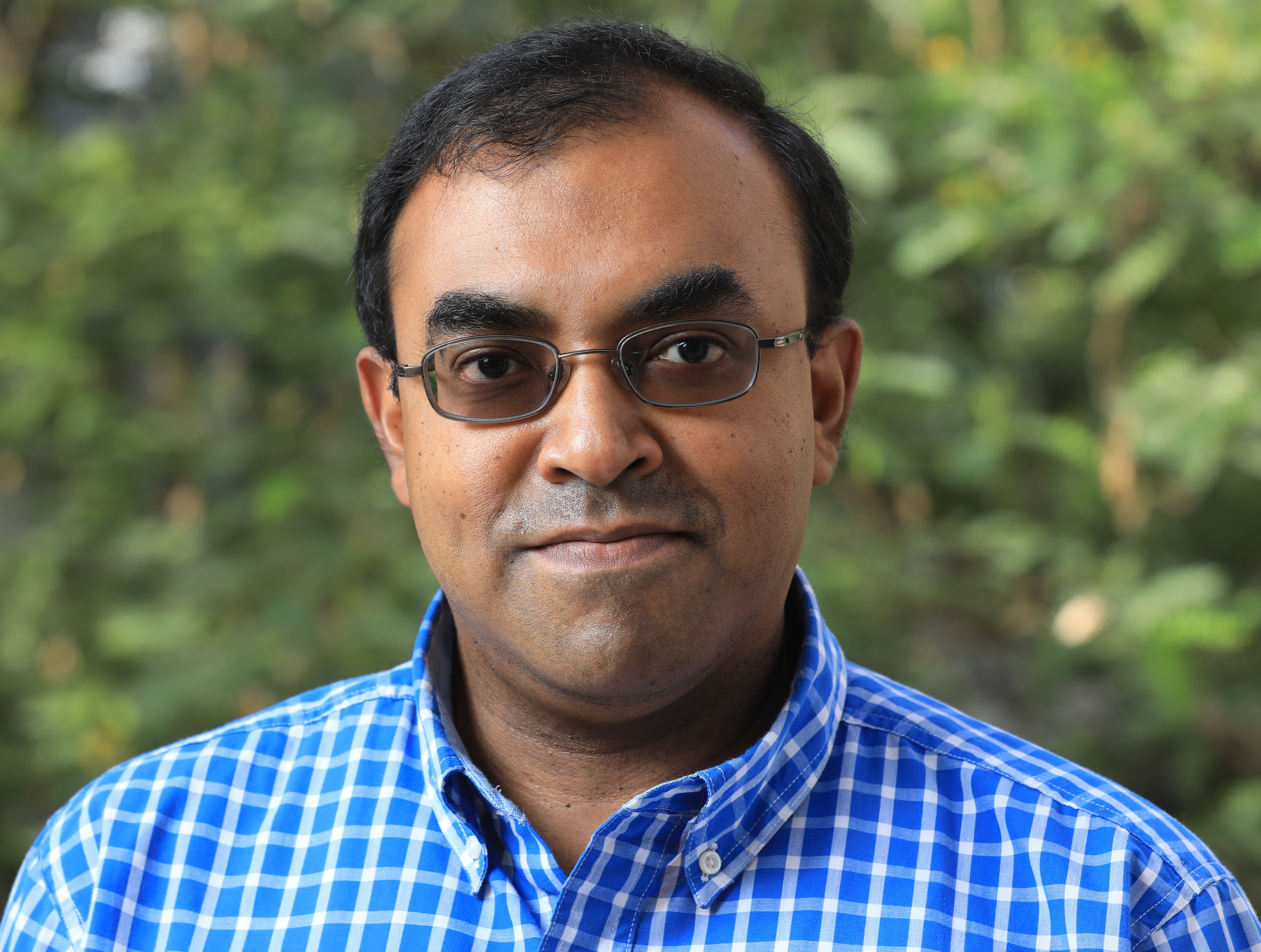}}]{Amit Roy-Chowdhury} received his PhD from the University of Maryland, College Park (UMCP) in Electrical and Computer Engineering in 2002 and joined the University of California, Riverside (UCR) in 2004 where he is a Professor and Bourns Family Faculty Fellow of Electrical and Computer Engineering, Director of the Center for Robotics and Intelligent Systems, and Cooperating Faculty in the department of Computer Science and Engineering. He leads the Video Computing Group at UCR, working on foundational principles of computer vision, image processing, and vision-based statistical learning, with applications in cyber-physical, autonomous and intelligent systems. He has published about 200 papers in peer-reviewed journals and conferences. He is the first author of the book Camera Networks: The Acquisition and Analysis of Videos Over Wide Areas. His work on face recognition in art was featured widely in the news media, including a PBS/National Geographic documentary and in The Economist. He is on the editorial boards of major journals and program committees of the main conferences in his area. His students have been first authors on multiple papers that received Best Paper Awards at major international conferences, including ICASSP and ICMR. He is a Fellow of the IEEE and IAPR, received the Doctoral Dissertation Advising/Mentoring Award 2019 from UCR, and the ECE Distinguished Alumni Award from UMCP.
\end{IEEEbiography}




\end{document}

%% file: sections/N01_rv2_abstract.tex
Prior works on text-based video moment localization focus on temporally grounding the textual query in an untrimmed video. These works assume that the relevant video is already known and attempt to localize the moment on that relevant video only. Different from such works, we relax this assumption and address the task of localizing moments in a corpus of videos for a given sentence query. This task poses a unique challenge as the system is required to perform: (i) retrieval of the relevant video where only a segment of the video corresponds with the queried sentence, and (ii) temporal localization of moment in the relevant video based on sentence query. Towards overcoming this challenge, we propose Hierarchical Moment Alignment Network (HMAN) which learns an effective joint embedding space for moments and sentences. In addition to learning subtle differences between intra-video moments, HMAN focuses on distinguishing inter-video global semantic concepts based on sentence queries. Qualitative and quantitative results on three benchmark text-based video moment retrieval datasets - Charades-STA, DiDeMo, and ActivityNet Captions - demonstrate that our method achieves promising performance on the proposed task of temporal localization of moments in a corpus of videos.




%% file: sections/N02_rv2_introduction.tex
\section{Introduction} 
Localizing activity moments in long and untrimmed videos is a prominent video analysis problem. Early works on moment localization were mostly limited by the use of a predefined set of labels to describe an activity \cite{lei2018temporal, chao2018rethinking,singh2016multi,lin2017single}. However, due to the nature of the complexity of real-life activities, natural language sentences would be the appropriate choice to describe an activity rather than a predefined set of labels. Recently, there are several works \cite{gao2017tall, anne2017localizing, wu2018multi, liu2018attentive, chen2018temporally, ge2019mac, xu2019multilevel, zhang2019man, yuan2019semantic, lin2020moment} that utilize sentence queries to temporally localize moments in untrimmed videos. All these approaches build upon an underlying assumption that the correspondence between sentences and videos is known. As a result, these approaches attempt to localize moments only in the related video. \textcolor{black}{We argue that such an assumption of knowing relevant videos a priori may not be plausible for most practical scenarios.} It is more likely that a user would need to retrieve a moment from a large corpus of videos given a sentence query.

\begin{figure}
  \includegraphics[width = \linewidth]{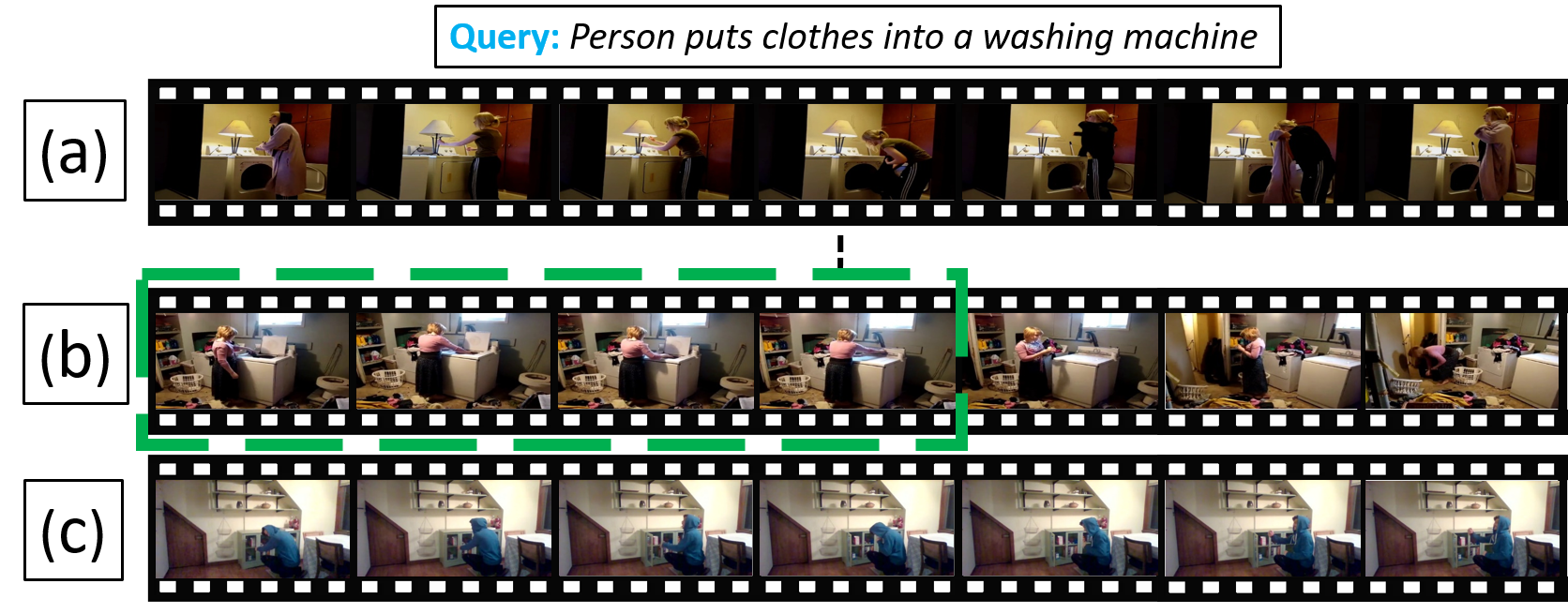}
  \caption{Example illustration of our proposed task. We consider localizing moments in a corpus of videos given a text query. Here, for the queried text: `Person puts clothes into a washing machine', the system is required to identify the relevant video-\textit{(b)} from the illustrated corpus of videos (video-\textit{(a)}, video-\textit{(b)}, and video-\textit{(c)}) and temporally localize the pertinent moment (ground truth moment marked by the green dashed box) in that relevant video.}
  \label{fig:prob_statement}
\end{figure}

In this work, we relax the assumption of specified video-sentence correspondence of the prior works on temporal moment localization and address the more challenging task of localizing moments in a corpus of videos. \update{For example in Figure \ref{fig:prob_statement}, the moment marked by the green dashed box in video-\textit{(b)} corresponds to the text query: \textit{`Person puts clothes into a washing machine'}. Prior works on temporal moment localization only attempt to detect the temporal endpoints in the given video-\textit{(b)} by learning to identify subtle changes in dynamics of the activity. However, the task of localizing the correct moment in the illustrated collection of videos (i.e., \textit{(a)}, \textit{(b)}, and \textit{(c)} in Figure \ref{fig:prob_statement}) imposes the additional requirement to distinguish moments from different videos and identify the correct video (video-\textit{(b)}) based on the differences of putting and pulling activities as well as the presence of washing machine and clothes.} 

To address this problem, a trivial approach would be to use an off-the-shelf video-text retrieval module to retrieve the relevant video and then localize the moment in that retrieved video. Most of the video-text retrieval approaches \cite{zhang2018cross, mithun2019joint, shao2018find, wray2019fine, dong2019dual, qi2021semantics,  wu2018unsupervised, feng2020exploiting} are designed for cases where videos and text queries have a one-to-one correspondence, i.e., a query sentence reflects a trimmed and short video or a query paragraph represents a long and untrimmed video. However, in our addressed task, the query sentence reflects a segment of a long and untrimmed video, and different segments of a video can be associated with different language annotations, resulting in one-to-many video-text correspondence. 
Hence, the existing video-text retrieval approaches are likely to fall short on our target task. Another trivial approach would be to scale up the temporal localization of moments approaches, i.e., instead of searching over a given video, it searches over the corpus of videos. \update{However, these approaches are only designed to discern intra-video moments based on sentence semantics and fail to distinguish moments from different videos and identify the correct video.}

In this work, based on the text query, we focus on discerning moments from different videos as well as understand the nuances of activities simultaneously to localize the correct moment in the relevant video. Our objective is to learn a joint embedding space that will align representations of corresponding video moments and sentences. For this, we propose \textbf{H}ierarchical \textbf{M}oment \textbf{A}lignment \textbf{N}etwork (\textbf{HMAN}), a novel neural network framework that effectively learns a joint embedding space to align corresponding video moments and sentences. Learning joint embedding space for retrieval or localization tasks has been addressed by several other methods \cite{anne2017localizing, escorcia2019temporal, feng2020exploiting, pan2016jointly, ye2020augmentation, ye2020probabilistic}. Among them, \cite{anne2017localizing} and \cite{escorcia2019temporal} are closely related to our work as they try to align corresponding moment and sentence representations in the joint embedding space. However, our approach is significantly different from these works. In contrast to these works, HMAN utilizes temporal convolutional layers in a hierarchical structure to represent candidate video moments. It allows the model to generate all candidate moment representations of a video in a single pass, which is more efficient than sliding based approaches like \cite{anne2017localizing, escorcia2019temporal}. Our learning objective is also different from \cite{anne2017localizing, escorcia2019temporal}, where they only try to distinguish between intra-video moments and inter-video moments. In our proposed approach, in addition to distinguishing intra-video moments, we propose a novel learning objective that utilizes text-guided global semantics to distinguish different videos. Global semantics of a video refers to the semantics that is common across most of the moments of that video. As the global semantics vary across videos, by distinguishing videos, we learn to distinguish inter-video moments. We demonstrate the advantage of our proposed approach over other baseline approaches and contemporary works on three benchmark datasets.

\subsection{Contributions}
The main contributions of the proposed work are as follows:

\begin{itemize}[leftmargin = *]
    \item We explore an important, yet under-explored, problem of text query-based localization of moments in a video corpus.
    \item We propose a novel framework, HMAN, that uses stacked temporal convolutional layers in a hierarchical structure to represent video moments and texts jointly in an embedding space. \update{Combined with the proposed learning objective, the model is able to align moment and sentence representations by distinguishing both local subtle differences of the moments as well as global semantics of the videos simultaneously.}
    \item \update{Towards solving the problem, we propose a novel learning objective that utilizes text-guided global semantics of the videos to distinguish moments from different videos.}
    \item We empirically show the efficacy of our proposed approach on DiDeMo, Charades-STA, and ActivityNet Captions dataset and study the significance of our proposed learning objective.

\end{itemize}

\begin{figure*}
    \centering
    \includegraphics[width= \textwidth]{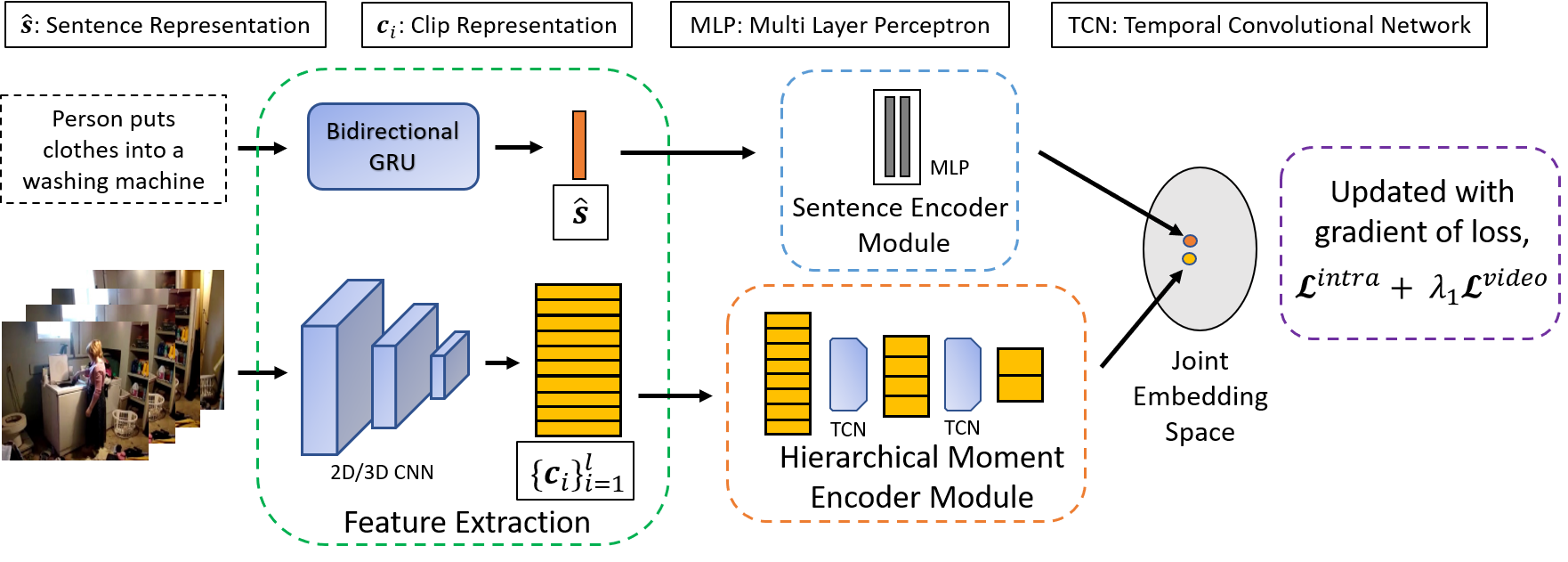}
    \caption{A brief illustration of the proposed Hierarchical Moment Alignment Network for the moment localization task in a video corpus. The framework uses the feature extraction unit to extract clip and sentence features. Hierarchical moment encoder module and sentence encoder module projects moment representations and sentence representations in the joint embedding space respectively. The network learns to align moment-sentence pairs in the joint embedding space by explicitly focusing on distinguishing intra-video moments and inter-video global semantic differences. (Details of the learning procedure in section \ref{learning})}
    \label{fig:concept}
\end{figure*}

%% file: sections/N03_rv2_related_work.tex
\section{Related Works}

\underline{\textit{Video-Text Retrieval.}} Among the cross-modal retrieval tasks \cite{lee2018stacked, liu2016sequential, deng2018triplet, dong2018predicting, mithun2018learning}, video-text retrieval has gained much attention recently. Emergence of datasets like the Microsoft Research Video to Text (MSR-VTT) \cite{xu2016msr}, the MPII movie description dataset as part of the Large Scale Movie Description Challenge (LSMDC) dataset \cite{rohrbach2015dataset}, and Microsoft Video Description Dataset (MSVD) \cite{chen2011collecting} have boosted video-text retrieval task. These datasets contain short video clips with accompanying natural language. Initial approaches for the video-text retrieval task were based on concept classification \cite{markatopoulou2017query, le2016nii, ueki2017waseda}. Recent approaches focus on directly encoding video and text in a common space and retrieving relevant instances based on some similarity measure in the common space \cite{dong2018predicting, mithun2018learning, xu2015jointly, yu2018joint, li2019w2vv++,chen2019weakly}. These works used Convolutional Neural Network (CNN) \cite{yu2018joint} or Long Short-Term Memory Network (LSTM) \cite{yu2017end} for video encoding. To encode text representations, Recurrent Neural Network (RNN) \cite{xu2015jointly}, bidirectional LSTM \cite{yu2018joint} and GRU \cite{mithun2019joint} were commonly used. Mithun et al. \cite{mithun2019joint} employed multimodal cues such as image, motion, and audio for video encoding. In \cite{dong2019dual}, multi-level encodings for video and text were used and both videos and sentences were encoded in a similar manner. Liu et al. \cite{liu2019use} proposed collaborative experts model to aggregate information effectively from different pre-trained experts. Yu et al. \cite{yu2018joint} proposed a Joint Sequence Fusion model for sequential interaction of videos and texts. Song et al. \cite{song2019polysemous} introduced Polysemous Instance Embedding Networks that compute multiple and diverse representations of an instance. Among the recent works, Wray et al. \cite{wray2019fine} enriched the embedding learning by disentangling parts-of-speech of captions. Chen et al. \cite{chen2020fine} used Hierarchical Graph Reasoning to improve fine-grained video-text retrieval. Another line of work considers video-paragraph retrieval. For example, Zhang et al. \cite{zhang2018cross} proposed hierarchical modeling of videos, and paragraphs and Shao et al. \cite{shao2018find} utilized top-level and part-level association for the task of video-paragraph retrieval. However, all of these approaches have an underlying assumption that videos and text queries have one-to-one correspondence. As a result, they are not adaptable for our addressed task, where the video-text pairs have one-to-many correspondence.

\underline{\textit{Temporal Localization of Moments.}} The task of localizing a moment/activity in a given long and untrimmed video via text query was introduced in \cite{gao2017tall, anne2017localizing}. After that, there have been a lot of works \cite{wu2018multi, liu2018attentive, chen2018temporally, ge2019mac, xu2019multilevel, zhang2019man, yuan2019semantic, jiang2019cross, liu2018cross, zhang2019exploiting, mithuncvpr2019, ghosh2019excl, yuan2019find, zhang2019cross, zhang2019learning, he2019read, hahn2019tripping, hendricks2018emnlp, liu2018temporal, regneri2013grounding} that addressed this task. All of these works on temporal localization of moments can be divided into two categories: i) two stage approaches that sample segments of videos in the first step and then try to find a semantic alignment between sentences and those segments of videos in the second step \cite{gao2017tall, anne2017localizing, wu2018multi, liu2018attentive, chen2018temporally, ge2019mac, xu2019multilevel}, and ii) single stage approaches that predict the association of sentences with multi-scale visual representation units as well as predict temporal boundary for each visual representation unit in a single pass \cite{zhang2019man, yuan2019semantic}. Among all the approaches, Gao et al. \cite{gao2017tall} developed Cross-modal Temporal Regression Localizer that jointly models text queries and video clips. A common embedding space for video temporal context features and language features was learnt in \cite{anne2017localizing}. Some of the works focused on vision-language fusion techniques to improve localization performance. For example, Multimodal Circulant Fusion was incorporated in \cite{wu2018multi}. Liu et al. \cite{liu2018attentive} incorporated a memory attention mechanism to emphasize the visual features mentioned in the query and simultaneously use their context. Ge et al. \cite{ge2019mac} mined activity concepts from both video and language modalities to improve the regression performance. Chen et al. \cite{chen2018temporally} proposed Temporal GroundNet which captures evolving fine-grained frame-by-word interactions. Xu et al. \cite{xu2019multilevel} used early integration of vision and language for proposal generation and query sentence modulation using visual features. Among the single shot approaches, candidate moment encoding and temporal structural reasoning were unified in a single shot framework  in \cite{zhang2019man}. Semantic Conditioned Dynamic Modulation (SCDM) was proposed in \cite{yuan2019semantic} for correlating sentence and related video contents. These approaches on moment localization in a given video show promise, but fall short on realizing the requirement of identifying the correct video to address the task of moment localization in a corpus of videos.

There has been one concurrent work \cite{escorcia2019temporal} that addressed the task of temporal localization of moments in a video corpus. They adopted the approach of Moment Context Network \cite{anne2017localizing}. However, instead of directly learning moment-sentence alignment as in \cite{anne2017localizing}, they tried to learn clip-sentence alignment for scalability issues where a moment consists of multiple clips. Even so, a referring event is likely to consist of multiple clips, and a single clip can not reflect the complete dynamics of an event. Hence, consecutive clips with different contents need to be aligned with the same sentence which results in suboptimal representation for both the clips and the sentence. 
We later empirically show that our approach is significantly more effective than \cite{escorcia2019temporal} in the addressed task.

%% file: sections/N04_rv2_methodology.tex
\section{Methodology}
In this section, we present our framework for the task of text-based temporal localization of moments in a corpus of untrimmed and unsegmented videos. First, we define the problem and provide an overview of the HMAN framework. Then, we present how clip-level video representations and word-level sentence representations are extracted. Then, we describe the framework in detail along with the hierarchical temporal convolutional network to generate moment embeddings and sentence embeddings. Finally, we describe how we learn to encode moment and sentence representations in the joint embedding space for effective retrieval of the moment based on a text query.

\begin{figure*}
    \centering
    \includegraphics[width =  \linewidth]{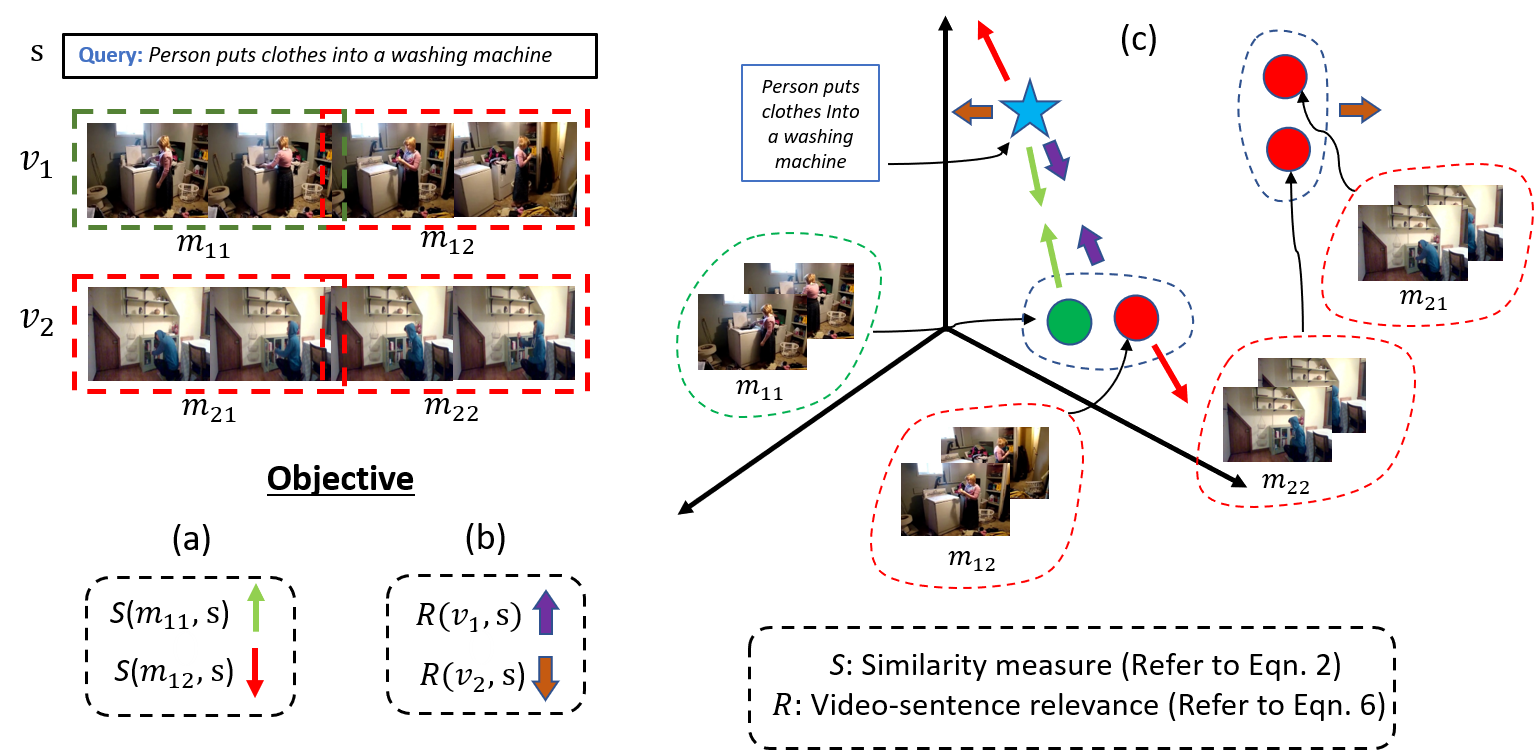}
    \caption{A conceptual representation of our proposed learning objective. For a text query $s$ with relevant moment $m_{11}$ in a set of videos $\{v_1,v_2\}$ with set of moments $\{m_{11},m_{12},m_{21},m_{22}\}$, we learn the joint embedding space using- (a) intra-video moments: increasing similarity for relevant pair $(m_{11},s)$ and decreasing similarity for non-relevant pair $(m_{12},s)$ from the same video, (b) global semantics of video: increasing video-sentence relevance for relevant pair $(v_1,s)$ and decreasing for non-relevant pair $(v_2,s)$, where the video-sentence relevance is computed in terms of moment-sentence similarity. This is also illustrated in (c), where the arrows indicate which pairs are learning to increase their similarity (moving close in the embedding space) and which pairs are learning to decrease their similarity (moving further away in the embedding space). Details can be found in section \ref{learning} }
    \label{fig:loss}
\end{figure*}

\subsection{Problem Statement} Consider that we have a set of $N$ long and untrimmed videos $\mathcal{V} = \{v_i\}_{i=1}^N$, where a video $v$ is associated with $m_v$ temporal sentence annotations $\mathcal{T}=\{(s_j,\tau_j^s,\tau_j^e)\}_{j=1}^{m_v}$. Here, $s_j$ is the sentence annotation and $\tau_j^s,\tau_j^e$ are the starting time and ending time of the moment in the video that corresponds with the sentence annotation $s_j$. The set of all temporal sentence annotations is $\mathcal{S}=\{\mathcal{T}_i\}_{i=1}^{N}$. Given a natural language query $s$, our task is to predict a set $s_{det}=\{v,\tau^s,\tau^e \}$ where, $v$ is the video that contains the relevant moment and $\tau^s, \tau^e$ are the temporal information of that moment. 

\vspace{-4mm}

\subsection{Framework Overview}
Our goal is to learn representations for candidate moments and sentences in such a way that the related moment-sentence pairs are aligned in the joint embedding space. Towards this goal, we propose HMAN, which is illustrated in Figure \ref{fig:concept}. First, we employ a feature extraction unit to extract clip level features $\{\boldsymbol{c}_i\}_{i=1}^{l}$ from a video and sentence features $\hat{\boldsymbol{s}}$ from a sentence. Clip representations and sentence representations are used to learn the semantic alignment between sentences and candidate moments. To project the moment representations and sentence representations in the joint embedding space, we use a hierarchical moment encoder module and a sentence encoder module respectively. The moment encoder module is inspired by single shot temporal action detection approach \cite{lin2017single} where temporal convolutional layers are stacked in a hierarchical structure to obtain multi-scale moment features representing video segments of different duration. For the sentence encoder module, we use a two-layer feedforward neural network. Based on text queries, we derive the learning objective to explicitly focus on distinguishing intra-video moments and inter-video global semantics. We adopted sum-margin based triplet loss \cite{frome2013devise} and max-margin based triplet loss \cite{frome2013devise} separately in two different settings to train the model in an end-to-end fashion and gained performance improvement over baseline approaches in both setups. In the inference stage, for a query sentence, the candidate moment with the most similar representation is retrieved from the corpus of videos.

\subsection{Feature Extraction Unit}
To work with data from different modalities, we extract feature representations using modality specific pretrained models.

\underline{\textit{Video Feature Extraction.}} We extract high level video features using a deep convolutional neural network. Each video $v$ is divided into a set of $l$ non-overlapping clips and we extract features for each clip. As a result, the video is represented by a set of features $\{\boldsymbol{c}_i\}_{i=1}^l$, where $\boldsymbol{c}_i$ is the feature representation of the $i^{th}$ clip. To generate representations for all the candidate moments of a video in a single shot approach \cite{lin2017single}, we keep the input video length, i.e., number of clips, $l$, fixed. A video longer than the fixed length is truncated and a video shorter than the fixed length is padded with zeros.     

\underline{\textit{Sentence Feature Extraction.}} To represent sentences, we use GloVe word embedding \cite{pennington2014glove} for each word in a sentence. Then these word embedding sequences are encoded using a Bi-directional Gated Recurrent Unit (GRU) \cite{chung2014empirical} with $512$ hidden states. Here, words in a sentence are represented by a $512$-dimensional vector, corresponding to their GRU hidden states. So, we can have a set of word-by-word representations of a sentence $S=\{\boldsymbol{h}_i\}_{i=1}^n$, where $n$ is the number of words present in the sentence. The average of the word representations is used as the sentence representation $\boldsymbol{\hat{s}}$. 

\vspace{-1mm}
\subsection{Moment Encoder Module} 
Existing approaches for moment localization based on learning joint visual-semantic embedding space either use a temporal sliding window with multiple scales \cite{anne2017localizing} or optimize over a predefined set of consecutive clips based on clip-sentence similarity \cite{escorcia2019temporal} to generate candidate segments. However, sliding over a video with different scales or optimizing for all possible combinations of clips is computationally heavy. Again, in both cases, extracted candidate moments or predefined clips are projected in the joint embedding space independent of neighboring or overlapping moments/clips of the same video. Consequently, while learning the moment-sentence or clip-sentence semantic alignment, representations for neighboring or overlapping moments are not constrained to be well clustered to preserve the semantic similarity. Therefore, instead of projecting representations for candidate moments independently and inefficiently in the joint embedding space, \aqe{inspired by the single shot activity detection \cite{lin2017single}, we use temporal convolutional layers \cite{lea2016temporal} in a hierarchical setup} to project representations for all candidate moments of a video simultaneously. We use a stack of $1D$ convolutional layers where the convolution operation can be denoted as $Conv(\sigma_k,\sigma_s,d)$. Here, $\sigma_k$, $\sigma_s$, and $d$ indicate the kernel size, stride size, and filter numbers, respectively. The set of moment representations generated for $K$ layers of hierarchical structure is $\{\{ \boldsymbol{m}_i^k\}_{i=1}^{T_k}\}_{k=1}^{K}$. Here, $T_k$ is the temporal dimension of the $k^{th}$ layer, which decreases in the following layers. $\boldsymbol{m}^k_i \in \mathcal{R}^{d}$ is the $i^{th}$ moment representation of the $k^{th}$ layer and $k^{th}$ layer generates $T_k$ moment representations. Feature representations in the top layers of the hierarchy correspond to moments with shorter temporal duration, while the feature representations in the bottom layers correspond to moments with longer duration in a video. \aqe{We keep the feature dimension of each moment representation fixed to $d$ for all the layers of the temporal convolutional network.}


\subsection{Sentence Encoder Module}

We learn to project the textual representations in the joint embedding space keeping the inputs from different modalities with similar semantics close to each other. We use two layers of feedforward neural network with learnable parameters $\boldsymbol{W}^s_1$, $\boldsymbol{W}^s_2$, $\boldsymbol{b}^s_1$, and $\boldsymbol{b}^s_2$ to project the sentence representation $\boldsymbol{\hat{s}}$ in the joint embedding space, which can be defined as, 
\begin{equation}
    \boldsymbol{s} =  \boldsymbol{W}^s_2   \Big(BN\big(ReLU(\boldsymbol{W}^s_1\hat{\boldsymbol{s}}+\boldsymbol{b}^s_1)\big)\Big)+\boldsymbol{b}^s_2
\end{equation}
Here, the dimension of the projected sentence representation $\boldsymbol{s}$ is kept consistent with the projected moment representation $\boldsymbol{m}$ ($\boldsymbol{m},\boldsymbol{s} \in \mathcal{R}^d$).

\begin{algorithm}[t]
    \caption{Learning optimized HMAN (max-margin case)}
	\label{algo:hman} 
	\begin{algorithmic}
		\State {\bf Input:} Untrimmed video set $\mathcal{V}$, Temporal sentence annotation set $\mathcal{S}$, Initialized HMAN weights $\theta$	
		\For{t = 1 to maxIter}		
		\State \textbf{step 1:} Construct minibatch of video-sentence pairs
		\State \textbf{step 2:} Extract video and sentence feature 
		\State \textbf{step 3:} Project candidate moment and sentence \\ \ \ \ \ representations in the joint embedding space	
		\State \textbf{step 4:} Construct triplets 
		\State \textbf{step 5:} Compute  $\mathcal{L}^{intra}_{max}$ and $\mathcal{L}^{video}_{max}$ using Eqn. \ref{eqn:intra_max} \& \ref{eqn:video_max}
		\State \textbf{step 6:} Optimize $\theta$ by minimizing total loss
		\EndFor
		\State {\bf Output:}
		Optimized HMAN weights $\theta$
    \end{algorithmic}
\end{algorithm}

\subsection{\update{Learning Joint Embedding Space}}
\label{learning}

Projected representations in the joint embedding space from different modalities need to be close to each other if they are semantically related. Training procedures to learn projecting representations in the joint embedding space mostly adopts two common loss functions: sum-margin based triplet ranking loss \cite{frome2013devise} and max-margin based triplet ranking loss \cite{faghri2017vse++}. We consider both of these loss functions separately. As illustrated in Figure \ref{fig:loss}, we focus on distinguishing intra-video moments and inter-video global semantic concepts. In this section, we discuss our approach to learn projecting representations from different modalities in the joint embedding space for multimodal data.

\underline{\textit{Similarity Measure.}} 
We use the cosine similarity of projected representations from two modalities in the joint embedding space to infer their semantic relatedness. So, the similarity between a candidate moment $m$ and a sentence $s$ is, 
\begin{equation}
    S(m,s) = \frac{\boldsymbol{m}^T\boldsymbol{s}}{ \|\boldsymbol{m}\| \|\boldsymbol{s}\| }
    \label{eqn:sim}
\end{equation}
where $\boldsymbol{m}$ and $\boldsymbol{s}$ are the projected moment representation and sentence representation in the joint embedding space.

\update{\underline{\textit{Learning for Intra-video Moments.}}} To localize a sentence query in a video, the model needs to identify the subtle differences of the candidate moments from the same video and distinguish them. Among the candidate segments of a video, one or few of the moments can be considered related to the query sentence based on some IoU threshold. While training the network, we consider related moments with the queried sentence as the positive pairs and non-corresponding moments with the queried sentence as the negative pairs. Suppose, for a pair of video-sentence $(v,s)$, we consider the set of positive moment-sentence pairs $\{(m,s)\}$ and the set of negative moment-sentence pairs $\{(m^-,s)\}$. We compute the intra-video ranking loss for all video-sentence pairs $\{(v,s)\}$. Using the sum-margin setup, the intra-video triplet loss is:   

\vspace{-.2cm}
\begin{equation}
     \mathcal{L}^{intra}_{sum} = \sum_{\{(v,s)\}} \sum_{ \{(m,s)\} }\sum_{\{(m^-,s)\}}\big[\alpha_{intra}-S(m,s)+S(m^-,s)\big]_+
\end{equation}

Similarly, using the max-margin setup, we calculate the intra-video triplet loss by, 

\begin{equation}
    \hat{m} = \argmax_{m^-} S(m^-,s)
\end{equation}

\begin{equation}
     \mathcal{L}^{intra}_{max} = \sum_{\{(v,s)\}} \sum_{ \{(m,s)\} }\big[\alpha_{intra}-S(m,s)+S(\hat{m},s)\big]_+
     \label{eqn:intra_max}
\end{equation}
Here, $[f]_+ = max(0,f)$ and $\alpha_{intra}$ is the ranking loss margin for intra-video moments.

\update{\underline{\textit{Learning for Videos.}}} Learning to distinguish intra-video moments only allows the model to learn subtle changes in the video. It does not allow the model to distinguish moments from different videos. However, learning to differentiate moments from different videos is important as we need to localize the correct moment in the video corpus. Hence, we also learn to distinguish moments from different videos by capitalizing on the text-guided global semantics of videos. As the global semantics varies across videos we try to distinguish videos based on these global semantics. To do so, we learn to maximize the relevance of correct video-sentence pairs. Video-sentence relevance is computed in terms of moment-sentence relevance. As a result, learning to align video-sentence pairs enforces constraints on the representation of moments from different videos to be dissimilar. Inspired by the work of \cite{lee2018stacked}, we compute the relevance of a video and a sentence by,
\begin{equation}
    R(v,s) = \log \Big( \sum_{\{m\}} \exp \big(\beta S(m,s)\big) \Big)^{\sfrac{1}{\beta}},
    \label{eqn:relevence}
\end{equation}
where $\beta$ is a factor that determines how much to magnify the importance of the most relevant moment-sentence pair and $\{m\}$ is the set of all the moments in video $v$. 
As $\beta \rightarrow \infty$, $R(v,s)$ approximates $\max_{m_i \in v} S(m_i,s)$. This is necessary because all the segments of the video do not correspond to the sentence. 

For each positive video-sentence pair $(v,s)$ where the sentence $s$ relates to a segment of the video $v$, we can consider two sets of negative pairs $\{(v^-, s)\}$ and $\{(v,s^-)\}$. Using the sum-margin setup, we calculate the triplet loss for video-sentence alignment of all the positive video-sentence pairs $\{(v,s)\}$ by,
\begin{align}
     \mathcal{L}^{video}_{sum} & = \sum_{\{(v,s)\}} \sum_{\{(v^-,s)\}} \big[\alpha_{video}-R(v,s)+R(v^-,s)\big]_+ \nonumber \\
     & + \sum_{\{(v,s)\}} \sum_{\{(v,s^-)\}} \big[\alpha_{video}-R(v,s)+R(v,
     s^-)\big]_+
\end{align}

Similarly, using the max-margin setup, we compute the triplet loss for video-sentence alignment by,
\begin{equation}
    \hat{v} = \argmax_{v^-} R(v^-,s)
\end{equation} 
\begin{equation}
    \hat{s} = \argmax_{s^-} R(v,s^-)
\end{equation}
\begin{align}
     \mathcal{L}^{video}_{max} = & \sum_{\{(v,s)\}} \big[\alpha_{video}-R(v,s)+R(\hat{v},s)\big]_+ \nonumber \\
     & + \sum_{\{(v,s)\}} \big[\alpha_{video}-R(v,s)+R(v,
     \hat{s})\big]_+
     \label{eqn:video_max}
\end{align}
Here, $\alpha_{video}$ is the ranking loss margin for learning inter-video global semantic concepts.

\begin{table}[t]
\centering
  \caption{Tabulated summary of the details of dataset contents}
  \label{tab:dataset}
  \resizebox{\linewidth}{!}{
  
  \begin{tabular}{l|c|c|c}
    \toprule
    & \multicolumn{2}{c|}{\underline{Number of videos}} & Moment-sentence\\
    Dataset & Total & Train/Val/Test & pairs\\
    \midrule
    DiDeMo & 10464 & 8395 / 1065 / 1004 & 26892\\
    Charades-STA & 6670 & 5336 / - / 1334 & 16128 \\
    ActivityNet Captions & 20k & 10009 / 4917 / - & 71942\\
    \bottomrule
  \end{tabular}
  }
\end{table}

\update{\underline{\textit{Learning Objective.}}} We combine the calculated loss for intra-video case and video-sentence alignment case and try to minimize it as our final objective. For the sum-margin setup, the final objective is,
\begin{equation}
     \min_{\theta}  \mathcal{L}^{intra}_{sum} + \lambda_1 \mathcal{L}^{video}_{sum}+ \alpha \|\mathcal{W}\|_{\mathit{F}}^2
     \label{eqn:total_loss_sum}
\end{equation}

Similarly, for the max-margin setup, the final objective is,
\begin{equation}
     \min_{\theta}  \mathcal{L}^{intra}_{max} + \lambda_1 \mathcal{L}^{video}_{max}+ \alpha \|\mathcal{W}\|_{\mathit{F}}^2
     \label{eqn:total_loss_max}
\end{equation}

Here, $\theta$ represents the network weights and all the learnable weights are lumped together in $\mathcal{W}$. $\lambda_1$ balances the contribution between learning to distinguish intra-video moments and learning to distinguish videos based on a text query. $\alpha$ is the weight on the regularization loss. Our objective is to optimize $\theta$ to generate a proper representation for candidate moments and sentences to minimize these combined losses. During training, these losses are computed for a mini-batch where the mini-batches are sampled randomly from the entire training set. This stochastic approach yields the advantage of reducing the probability of selecting instances with high semantic relation as the negative samples.

\underline{\textit{Inference.}} In the inference step, for a query sentence, we compute the similarity of candidate moment representations with the query sentence representation using Eqn. \ref{eqn:sim}. We retrieve the candidate moment from the video corpus that results in the highest similarity.

%% file: sections/N05_rv2_experiment.tex
\section{Experiments}
In this section, we experimentally evaluate the performance of our proposed method for the task of temporal localization of moments in a corpus of video. We first discuss the datasets we use and the implementation details of the experiments. Then we report and analyze the results both quantitatively and qualitatively.

\begin{table}[t]
\centering
  \caption{Tabulated summary of the implementation details regarding video processing for three datasets}
  \label{tab:implementation}
  \resizebox{\linewidth}{!}{
  \begin{tabular}{l|c|c|c|c}
    \toprule
    & Video & \# of candidate & Per Unit & Temporal dimension \\
    Dataset & length & moments & duration & of layers\\
    \midrule
    DiDeMo & 12 & 21 & 2.5s & \{6,5,4,3,2,1\}\\
    Charades-STA & 64 & 61 & 1s & \{31,16,8,4,2,1\}\\
    ActivityNet Captions & 512 & 1023 & 1s & \{512, 256, 128, 64, 32,\\
    & & & &  16, 8, 4 ,2, 1\}\\
    \bottomrule
  \end{tabular}
  }
\end{table}

\begin{table*}[t]
\centering
  \caption{Comparison of performance for the task of temporally localizing moments in a video corpus on DiDeMo dataset. \scriptsize{(${\dagger}$ reported from \cite{escorcia2019temporal}) ($\downarrow$ indicates the performance is better if the score is low)}}
  \label{tab:main_didemo}
  \begin{tabular}{l|c|cccccc}
    \toprule
    &  & \multicolumn{6}{c}{\underline{DiDeMo}}\\
    \multirow{2}{*}{} & Feature used & \multicolumn{3}{c}{\underline{$IoU = 0.50$}} & \multicolumn{3}{c}{\underline{$IoU = 0.70$}}\\
    &  & R@10 & R@100 & MR$\downarrow$ & R@10 & R@100 & MR$\downarrow$\\
    \midrule
    Moment Prior$^{\dagger}$ \cite{escorcia2019temporal} & - & 0.22 & 2.34 & 2527 & 0.17 & 1.99 & 3234\\
    MCN$^{\dagger}$ \cite{anne2017localizing} & RGB (ResNet-152) & 2.15 & 12.47 & 1057 & 1.55 & 9.03 & 1423\\
    SCDM \cite{yuan2019semantic} & RGB (ResNet-152) + Flow (TSN) & 0.57 & 4.43 & - & 0.22 & 1.42 & -\\
    VSE++ \cite{faghri2017vse++} + SCDM \cite{yuan2019semantic} & RGB (ResNet-152) + Flow (TSN) & 0.70 & 4.16 & - & 0.30 & 2.81 & -\\
    CAL$^{\dagger}$ \cite{escorcia2019temporal} & RGB (ResNet-152) & 3.90 & 16.51 & 831 & 2.81 & 12.79 & 1148\\
    \midrule
    HMAN (sum-margin, Eqn. \ref{eqn:total_loss_sum}) & RGB (ResNet-152) & 5.63 & 26.49 & 412 & 4.51 & 20.82 & 546\\
    \midrule
    HMAN (TripSiam \cite{dong2018triplet}) & RGB (ResNet-152) + Flow (TSN) & 2.34 & 17.82 & 509 & 1.59 & 13.92 & 637\\
    HMAN (DSLT \cite{lu2020deep}) & RGB (ResNet-152) + Flow (TSN) & 5.95 & 25.45 & 313 & 4.66 & 20.04 & 447\\
    HMAN (sum-margin, Eqn. \ref{eqn:total_loss_sum}) & RGB (ResNet-152) + Flow (TSN) & \textbf{6.25} & \textbf{28.39} & \textbf{302} & \textbf{4.98} & \textbf{22.51} & \textbf{416}\\
    HMAN (max-margin, Eqn. \ref{eqn:total_loss_max}) & RGB (ResNet-152) + Flow (TSN) & 5.47 & 20.82 & 618 & 3.86 & 16.28 & 905\\
    \bottomrule
  \end{tabular}
\end{table*}

\begin{table*}[t]
\centering
  \caption{Comparison of performance for the task of temporally localizing moments in a video corpus on Charades-STA dataset. \scriptsize{(${\dagger}$ reported from \cite{escorcia2019temporal}) ($\downarrow$ indicates the performance is better if the score is low)}}
  \label{tab:main_charades}
  \begin{tabular}{l|c|cccccc}
    \toprule
    & & \multicolumn{6}{c}{\underline{Charades-STA}}\\
    \multirow{2}{*}{} & Feature used &
    \multicolumn{3}{c}{\underline{$IoU = 0.50$}} & \multicolumn{3}{c}{\underline{$IoU = 0.70$}}\\
    & & R@10 & R@100 & MR$\downarrow$ & R@10 & R@100 & MR$\downarrow$\\
    \midrule
    Moment Prior$^{\dagger}$ \cite{escorcia2019temporal} & - & 0.17 & 1.63 & 4906 & 0.05 & 0.56 & 11699\\
    MCN$^{\dagger}$ \cite{anne2017localizing} & RGB (ResNet-152) & 0.52 & 2.96 & 6540 & 0.31 & 1.75 & 10262\\
    SCDM \cite{yuan2019semantic} & RGB (I3D) & 0.73 & 6.41 & - & 0.56 & 4.23 & -\\
    VSE++ \cite{faghri2017vse++} + SCDM \cite{yuan2019semantic} & RGB (I3D) & 1.02 & 5.06 & - & 0.70 & 3.37 & -\\
    CAL$^{\dagger}$ \cite{escorcia2019temporal} & RGB (ResNet-152) & 0.75 & 4.39 & 5486 & 0.42 & 2.78 & 8627\\
    \midrule
    HMAN (TripSiam \cite{dong2018triplet}) & RGB (I3D) & 1.27 & 7.60 & 2821 & 0.70 & 4.49 & 5766 \\
    HMAN (DSLT \cite{lu2020deep}) & RGB (I3D) & 1.05 & 7.27 & 2390 & 0.54 & 4.61 & \textbf{5496} \\
    HMAN (sum-margin, Eqn. \ref{eqn:total_loss_sum}) & RGB (I3D) & 1.29 & 7.73 & 2418 & 0.83 & 4.12 & 6395 \\
    HMAN (max-margin, Eqn. \ref{eqn:total_loss_max}) & RGB (I3D) & \textbf{1.40} & \textbf{7.79} & \textbf{2183} & \textbf{1.05} & \textbf{4.69} & 5812\\
    \bottomrule
  \end{tabular}
\end{table*}

\subsection{Datasets}
We conduct experiments and evaluate the performance on three benchmark text-based video moment retrieval datasets, namely DiDeMo \cite{anne2017localizing}, Charades-STA \cite{gao2017tall}, and ActivityNet Captions \cite{krishna2017dense}. All of these datasets contain unsegmented and untrimmed videos with natural language sentence annotations with temporal information. Table \ref{tab:dataset} summarizes the details of the contents of three datasets. 

\underline{\textit{DiDeMo.}} The Distinct Describable Moments (DiDeMo)
dataset \cite{anne2017localizing} is one of the most diverse datasets for
the temporal localization of moments in videos given natural
language descriptions. The videos are collected from Flickr and each video is trimmed to a maximum of 30 seconds. The videos in the dataset are divided into 5-second segments to reduce the complexity of annotation. The dataset is split into training, validation, and test sets containing 8,395, 1,065, and 1,004 videos respectively. The dataset contains a total of 26,892 moment-sentence pairs and each natural language description is temporally grounded by multiple annotators. 

\underline{\textit{Charades-STA.}} Charades-STA dataset is introduced in \cite{gao2017tall} to address the task of temporal localization of moments in untrimmed videos. The dataset contains a total of $6{,}670$ videos with $16{,}128$ moment-sentence pairs. We have used the published split of videos during training and testing (train-$5{,}336$, test-$1{,}334$). As a result, the training set and the testing set contain $12{,}408$ and $3{,}720$ moment-sentence pairs respectively. This dataset is originally built on the Charades \cite{sigurdsson2016hollywood} activity dataset with temporal activity annotation and video-level description. Authors in \cite{gao2017tall} adopted a keyword matching strategy to generate clip-level sentence annotation. 

\vspace{-1mm}

\update{\underline{\textit{ActivityNet Captions.}}}
ActivityNet Captions \cite{krishna2017dense} dataset, which is proposed for dense video captioning task, is built on the ActivityNet dataset \cite{heilbron2015activitynet}. It consists of YouTube video footage where each video contains at least two ground truth segments and each segment is paired with one ground truth caption \cite{xu2019multilevel}. This dataset contains around 20k videos which are split into training, validation, and testing set. We use the published splits over videos (train set – $10{,}009$ videos, validation set – $4{,}917$ videos), where the evaluation is done on the validation set. Videos are typically longer in length than DiDeMo and Charades-STA datasets.

\subsection{Evaluation Metric}
We use the standard evaluation criteria adopted by various previous temporal moment localization works \cite{gao2017tall, yuan2019semantic, zhang2019man}. These works use $R@k, IoU{=}m$ metric, which reports the percentage of cases where at least one of the top-$k$ results have Intersection-over-Union $(IoU)$ larger than $m$ \cite{gao2017tall}. For a sentence query, $R@k, IoU{=}m$ reflects if one of the top-$k$ retrieved moments has Intersection-over-Union with the ground truth moment larger than the specified threshold $m$. So, for each query sentence, $R@k, IoU{=}m$ is either 1 or 0. As this metric is associated with a queried sentence, we compute it for all the sentence queries in the testing set (DiDeMo, Charades-STA) or in the validation set (ActivityNet Captions) and report the average results. We report $R@k, IoU{=}m$ over all queried sentences for $k \in \{10,100\}$ and $m \in \{0.50, 0.70\}$. We also use median retrieval rank (MR) as an evaluation metric. MR computes the median of the rank of the correct moment for each query. Lower values of MR indicate good performance. We compute MR for $IoU \in \{0.50, 0.70\}$. Note that DiDeMo dataset provides multiple temporal annotations for each sentence. We consider a detection is correct if it overlaps with a minimum of two temporal annotations with a specified $IoU$.

\begin{table}[t]
\centering
  \caption{ Comparison of performance for the task of temporally localizing moments in a video corpus on ActivityNet Captions dataset. \scriptsize{(${\dagger}$ reported from \cite{escorcia2019temporal})}}
  \label{tab:main_activitynet}
  \resizebox{\linewidth}{!}{
  \begin{tabular}{l|c|cccc}
    \toprule
    & & \multicolumn{4}{c}{\underline{ActivityNet Captions}}\\
    \multirow{2}{*}{} & Feature & \multicolumn{2}{c}{\underline{$IoU = 0.50$}} & \multicolumn{2}{c}{\underline{$IoU = 0.70$}}\\
    & used & R@10 & R@100 & R@10 & R@100\\
    \hline
    Moment Prior$^{\dagger}$ & - & 0.05 & 0.47 & 0.03 & 0.26\\
    MCN$^{\dagger}$ \cite{anne2017localizing} & RGB (ResNet-152) & 0.18 & 1.26 & 0.09 & 0.70\\
    CAL$^{\dagger}$ \cite{escorcia2019temporal} & RGB (ResNet-152) & 0.21 & 1.58 & 0.10 & 0.90\\
    \midrule
    HMAN (sum) & RGB (C3D) & 0.43 & 2.84 & 0.22 & 1.48 \\
    HMAN (max) & RGB (C3D) & \textbf{0.66} & \textbf{4.75} & \textbf{0.32} & \textbf{2.27}\\
    \bottomrule
  \end{tabular}
  }
\end{table}

\begin{table*}[t]
\centering
  \caption{Comparison of the performance of HMAN with/without the Hierarchical moment Encoder Module. The experiments are done for DiDeMo and Charades-STA datasets. \scriptsize{(${\dagger}$ reported from \cite{escorcia2019temporal}) ($\downarrow$ indicates the performance is better if the score is low)}}
  \label{tab:w_tcn}  
  \begin{tabular}{l|cccccc|cccccc}
    \toprule
    \multirow{3}{*}{} & \multicolumn{6}{c|}{\underline{DiDeMo}} & \multicolumn{6}{c}{\underline{Charades-STA}}\\
    & \multicolumn{3}{c}{\underline{$IoU=0.50$}} & \multicolumn{3}{c|}{\underline{$IoU=0.70$}} & \multicolumn{3}{c}{\underline{$IoU=0.50$}} & \multicolumn{3}{c}{\underline{$IoU=0.70$}}\\
    & R@10 & R@100 & MR$\downarrow$ & R@10 & R@100 & MR$\downarrow$ & R@10 & R@100 & MR$\downarrow$ & R@10 & R@100 & MR$\downarrow$\\
    \midrule
    HMAN (sum, w/o TCN) & 3.44 & 14.14 & 1168 & 2.14 & 9.91 & 1636 & 1.13 & 6.12 & 4170 & 0.43 & 4.09 & 8295\\
    HMAN (sum, w/ TCN) & \textbf{6.25} & \textbf{28.39} & \textbf{302} & \textbf{4.98} & \textbf{22.51} & \textbf{416} & \textbf{1.29} & \textbf{7.73} & \textbf{2418} & \textbf{0.83} & \textbf{4.12} & \textbf{6395}\\
    \midrule
    HMAN (max, w/o TCN) & 3.41 & 12.13 & 1603 & 1.99 & 8.96 & 2214 & 0.70 & 4.71 & 5800 & 0.46 & 3.13 & 10907\\
    HMAN (max, w/ TCN) & \textbf{5.47} & \textbf{20.82} & \textbf{618} & \textbf{3.86} & \textbf{16.28} & \textbf{905} & \textbf{1.40} & \textbf{7.79} & \textbf{2183} & \textbf{1.05} & \textbf{4.69} & \textbf{5812}\\
    \bottomrule
  \end{tabular}  
\end{table*}

\begin{table}[t]
\centering
  \caption{Ablation study for the effectiveness of learning embedding space utilizing different loss components as described in \ref{learning} for DiDeMo dataset using sum-margin set up.}
  \label{tab:embedding_didemo}
  \begin{tabular}{l|cc|cc}
    \toprule
    \multirow{2}{*}{} & \multicolumn{2}{c|}{\underline{$IoU = 0.50$}} & \multicolumn{2}{c}{\underline{$IoU = 0.70$}}\\
    & R@10 & R@100 & R@10 & R@100\\
    \hline
    HMAN (intra) & 0.57 & 6.00 & 0.52 & 4.71 \\
    \hline
    HMAN (video) & 1.77 & 10.03 & 0.30 & 2.34 \\
    \hline
    HMAN (proposed) & \textbf{6.25} & \textbf{28.39} & \textbf{4.98} & \textbf{22.51}\\
    \bottomrule
  \end{tabular}
\end{table}

\subsection{Implementation Details}
For DiDeMo dataset, we use ResNet-152 features \cite{he2016deep}, where pool5 features are extracted at $5$ fps over the video frames. Then the features are max-pooled over $2.5s$ clips. Also, we extract optical flow features from the penultimate layer from a competitive activity recognition model \cite{wang2016temporal}. We use Kinetics pretrained I3D network \cite{carreira2017quo} to extract per second clip features for the Charades-STA dataset. For ActivityNet Captions dataset, we use extracted C3D features \cite{tran2015learning}. We set the number of input clips of a video, $l=12$ for DiDeMo dataset, $l=64$ for Charades-STA dataset, and $l=512$ for ActivityNet Captions dataset. Here, per unit length of input video represents non-overlapping clip of $2.5s$ duration for DiDeMo and non-overlapping clip of $1s$ duration for both Charades-STA and ActivityNet Captions dataset. For DiDeMo dataset, we use a fully connected layer followed by max-pool to generate representations with temporal dimension $6$ for each video. Then we use $6$ temporal convolutional layers to generate representations with temporal dimensions of $\{6,5,4,3,2,1\}$ resulting in representations for $21$ candidate moments. Similarly for Charades-STA, we use a fully connected layer followed by max-pool to generate representations with temporal dimension $32$ for each video. Then we use $6$ temporal convolutional layers with the temporal dimension of $\{32,16,8,4,2,1\}$ where we use the $31$ candidate moment representations from the last $5$ layers. Additionally, we use a branch temporal convolutional layer to generate representations of $30$ overlapping candidate moments, each with $6s$ duration and $2s$ stride. Combining these, we consider $61$ candidate moments for each video of Charades-STA dataset. For ActivityNet Captions dataset, we use a feedforward network followed by $10$ convolutional layers to generate representations with temporal dimension of $\{512,256,128,64,32,16,8,4,2,1\}$, resulting in $1023$ candidate moment representations. Table \ref{tab:implementation} illustrates the implementation details for video processing for all three datasets. we consider sentences with maximum of 15 words in length. If a sentence contains more than 15 words, the tailing words are truncated.

\begin{table}[t]
\centering
  \caption{Performance comparison for the task of retrieving correct video based on sentence query on DiDeMo and Charades-STA dataset.}
  \label{tab:R_retrieval}
  \begin{tabularx}{\linewidth}{l|>{\centering\arraybackslash}X>{\centering\arraybackslash}X>{\centering\arraybackslash}X|>{\centering\arraybackslash}X>{\centering\arraybackslash}X>{\centering\arraybackslash}X}
    \toprule
    \multirow{2}{*}{} & \multicolumn{3}{c|}{\underline{DiDeMo}} & \multicolumn{3}{c}{\underline{Charades-STA}}\\
    & R@10 & R@100 & R@200 & R@10 & R@100 & R@200\\
    \hline
    VSE++ \cite{faghri2017vse++}  & 2.49 & 16.81 & 29.53 & 1.89 & 13.31 & 24.43\\
    \hline
    HMAN (max) & 12.43 & 42.43 & 58.22 & 2.26 & 15.87 & 27.26\\
    \hline
    HMAN (sum) & \textbf{15.36} & \textbf{55.23} & \textbf{69.12} & \textbf{2.45} &  \textbf{18.51} & \textbf{30.52} \\
    \bottomrule
  \end{tabularx}
\end{table}

The proposed network is implemented in TensorFlow and trained using a single RTX 2080 GPU. To train the HMAN network, we use mini-batches containing $64$ video-sentence pairs for DiDeMo and Charades-STA and $32$ video-sentence pairs for ActivityNet Captions. We use the learning rate with exponential decay initializing from $10^{-3}$ for all three datasets. ADAM optimizer is used to train the network. We use 0.9 as the exponential decay rate for the first moment estimates and 0.999 as the exponential decay rate for the second-moment estimates. \aqe{We set $\alpha_{intra}$ and $\alpha_{video}$ to 0.05 and 0.20, respectively for all three datasets.} $\lambda_1$ is empirically set to $5$, $1$, and $1.5$, respectively for DiDeMo, Charades-STA, and ActivityNet Captions. $\alpha$ is set to $5\times10^{-5}$ for all three datasets.   

\begin{table*}[t!]
\centering
  \caption{Comparison of the performance of proposed LogSumExp pooling and average pooling. We compare the performance for the task of temporal localization of moments in video corpus for DiDeMo and Charades-STA dataset.}
  \label{tab:R_type}
  \begin{tabular}{l|cccc|cccc}
    \toprule
    \multirow{3}{*}{} & \multicolumn{4}{c|}{\underline{DiDeMo}} & \multicolumn{4}{c}{\underline{Charades-STA}}\\
    & \multicolumn{2}{c}{\underline{$IoU=0.50$}} & \multicolumn{2}{c|}{\underline{$IoU=0.70$}} & \multicolumn{2}{c}{\underline{$IoU=0.50$}} & \multicolumn{2}{c}{\underline{$IoU=0.70$}}\\
    & R@10 & R@100 & R@10 & R@100 & R@10 & R@100 & R@10 & R@100\\
    \midrule
    HMAN (sum, ave) & 5.63 & 26.05 & 4.43 & 20.82 & 1.10 & 7.19 & 0.62 & \textbf{4.47}\\
    HMAN (sum, log) & \textbf{6.25} & \textbf{28.39} & \textbf{4.98} & \textbf{22.51} & \textbf{1.29} & \textbf{7.73} & \textbf{0.83} & 4.12\\
    \midrule
    HMAN (max, ave) & 5.27 & 17.65 & \textbf{4.01} & 13.60 & 0.75 & 7.00 & 0.51 & 4.53\\
    HMAN (max, log) & \textbf{5.47} & \textbf{20.82} & 3.86 & \textbf{16.28} & \textbf{1.40} & \textbf{7.79} & \textbf{1.05} & \textbf{4.69}\\
    
    \bottomrule
  \end{tabular}
\end{table*}

\begin{table}[t]
\centering
  \caption{Ablation Study of the performance of HMAN (sum-margin) for Different Visual Features for DiDeMo dataset.}
  \label{tab:feature_table}
  \begin{tabular}{l|cc|cc}
    \toprule
    \multirow{2}{*}{} & \multicolumn{2}{c|}{\underline{$IoU = 0.50$}} & \multicolumn{2}{c}{\underline{$IoU = 0.70$}}\\
    & R@10 & R@100 & R@10 & R@100\\
    \hline
    VGGNet & 2.61 & 16.36 & 1.79 & 12.82\\
    \hline
    VGGNet + Flow & 3.98 & 21.29 & 3.14 & 16.76\\
    \hline
    ResNet & 5.63 & 26.49 & 4.51 & 20.82\\
    \hline
    ResNet + Flow & \textbf{6.25} & \textbf{28.39} & \textbf{4.98} & \textbf{22.51}\\
    \bottomrule
  \end{tabular}
\end{table}

\subsection{\update{Result Analysis}}
We conduct the following experiments to evaluate the performance of our proposed method:

\begin{itemize}[leftmargin=*]
    \item Comparison of the performance of proposed HMAN for the task of temporal localization of moments in video corpus with different baseline approaches and a concurrent work.
    \item Evaluation of the effectiveness of utilizing hierarchical moment encoder module.
    \item Investigation of the impact of learning joint embedding space by utilizing different components of the loss function (learning for intra-video moments ($\mathcal{L}^{intra}$) and learning for videos ($\mathcal{L}^{video}$).   
    \item Evaluation of the effectiveness of utilizing global semantics to identify the correct video.  
    \item Analyzing the effectiveness of video relevance computation (Eqn. \ref{eqn:relevence}) for the task of temporal localization of moments in a video corpus.
    \item Studying the performance of proposed HMAN for  different visual features.
    \item Performance comparison of HMAN with decreasing number of test set moment-sentence pairs.
    \item Evaluation of the run time efficiency.
    \item{Analysis of the $\lambda_1$ parameter sensitivity.}
\end{itemize}

\underline{\textit{\update{Temporal Localization of Moments in Video Corpus.}}} Table \ref{tab:main_didemo}, Table \ref{tab:main_charades}, and Table \ref{tab:main_activitynet} illustrate the quantitative performance of our framework for the task of temporal localization of moments in the video corpus. The evaluation setup considers $IoU \in \{0.50, 0.70\}$ and for each $IoU$ threshold, we report $R@10$, $R@100$ and MR. For a query sentence, the task requires to search over all the videos and retrieve the relevant moment. For example, in the DiDeMo dataset, the test set consists of $1{,}004$ videos totaling $4{,}016$ moment-sentence pairs. Again, we consider $21$ candidate moments for each video. So, for each query sentence, we need to search over $21 \times 1{,}004 = 21{,}084$ moment instances and retrieve the correct moment. This is itself a difficult task and the addition of ambiguity of similar kinds of activities in different videos makes the problem even harder. We compare the proposed method with the following baselines:
\begin{itemize}[leftmargin=*]
    \item \textbf{Moment Frequency Prior:} We use Moment Frequency Prior baseline from \cite{anne2017localizing}, which selects moments that correspond to gifs most frequently described by the annotators.
    \item \textbf{MCN:} The Moment Context Network \cite{anne2017localizing} for temporal localization of moments in a given video is scaled up to search moment from the entire video corpus.
    \item \textbf{SCDM:}
    The state-of-the-art Semantic Conditioned Dynamic Modulation (SCDM) network  \cite{yuan2019semantic} for temporal localization of moments in a video is scaled up to search over the entire video corpus.  
    \item \textbf{VSE++ + SCDM:} We use joint embedding based retrieval approach (VSE++ \cite{faghri2017vse++}) combined with SCDM as a baseline. In this setup, the framework first retrieves a few relevant videos (top $5\%$) and then localize moments on those retrieved videos using SCDM approach.
    \item \textbf{CAL:} We compare with Clip Alignment of Language \cite{escorcia2019temporal}. It is a concurrent work that addresses the task of localizing moments in a video corpus by aligning clip representation with language representation in the embedding space.
\end{itemize}

\begin{table}[t]
\centering
  \caption{Ablation study of the performance of HMAN (sum-margin) when the number of test set data is decreased for DiDeMo dataset.}
  \label{tab:decreasing}
  \resizebox{\linewidth}{!}{
  \begin{tabular}{l|ccc|ccc}
    \toprule
    \multirow{2}{*}{} & \multicolumn{3}{c|}{\underline{$IoU = 0.50$}} & \multicolumn{3}{c}{\underline{$IoU = 0.70$}}\\
    & R@10 & R@100 & MR$\downarrow$ & R@10 & R@100 & MR$\downarrow$\\
    \midrule
    HMAN (100\%) & 6.25 & 28.39 & 302 & 4.98 & 22.51 & 416\\
    \hline
    HMAN (50\%) & 6.90 & 30.15 & 268 & 5.68 & 23.73 & 372\\
    \hline
    HMAN (25\%) & 8.74 & 34.93 & 193 & 7.06 & 27.62 & 269\\
    \hline
    HMAN (10\%) & 13.35 & 45.60 & 102 & 10.30 & 36.65 & 142\\
    
    \bottomrule
  \end{tabular}
  }
\end{table}

\noindent
Note that we do not compare with baselines that utilize temporal endpoint features from \cite{anne2017localizing}, as these directly correspond to dataset priors and do not reflect a model’s capability \cite{liu2018temporal}. 


We observe that MCN and CAL perform better than the state-of-the-art SCDM approach in DiDeMo dataset but perform poorly compared to the SCDM approach in Charades-STA dataset. This is due to the fact that the video contents and language queries differ a lot among different datasets \cite{zhang2019man}. MCN and CAL learn to distinguish both intra-video moments and inter-video moments locally while SCDM only learns to distinguish intra-video moments. As DiDeMo dataset contains diverse videos of different concepts and relatively less number of candidate moments, learning to differentiate inter-video moments locally improves performance significantly. However, learning to differentiate inter-video moments locally does not have much impact on Charades-STA dataset. This also indicates the importance of distinguishing moments from different videos based on global semantics for a diverse set of video datasets. We also observe that in some of the cases, VSE++ + SCDM scores drop compared to the SCDM approach. Since the performance of VSE++ + SCDM depends on retrieving correct video, the localization performance drops if the retrieval approach fails to retrieve correct videos with higher accuracy.

For HMAN, we report the performance for both sum-margin and max-margin based triplet loss setups. Additionally, for DiDeMo and Charades-STA dataset, we report the performance of HMAN for two different loss calculation setups: TripSiam \cite{dong2018triplet} and DSLT \cite{lu2020deep}. In Table \ref{tab:main_didemo}, Compared to baseline approaches, the performance of our proposed approach is better for all metrics and outperforms other approaches with a maximum of $11.88\%$ absolute improvement in DiDeMo dataset. We observe that the sum-margin based triplet loss setup outperforms the max-margin setup, while both of these setups perform better than other baselines in DiDeMo dataset. For a fair comparison with CAL and MCN, we report the performance of HMAN with the ResNet-152 feature computed from RGB frames only. This setup also outperforms CAL and MCN. We also conduct experiment incorporating temporal end point feature in HMAN for DiDeMo dataset. It results in $\sim 0.5\% - 1\%$ improvement over HMAN (sum-margin) in $R@k$ metrics. It indicates the bias in the dataset where different types of events are correlated with different time frames of the video. In Table \ref{tab:main_charades}, for the Charades-STA dataset, the performance of HMAN is better for all metrics and the max-margin based triplet loss setup outperforms other baseline approaches with a maximum of $3.4\%$ absolute improvement. In Table \ref{tab:main_activitynet}, for ActivityNet Captions dataset, the HMAN max-margin setup outperforms other baselines with a maximum of $3.17\%$ absolute improvement. We do not compute SCDM and VSE++ + SCDM baselines for ActivityNet Captions dataset. Moment representations in SCDM and VSE++ + SCDM approaches are conditioned on sentence queries. For each query sentence, we need to compute moment representations from all the videos, resulting in $O(n^2)$ complexity. So testing on a set of $34{,}160$ query sentences and $4{,}917\times 1{,}023 = 5{,}030{,}091$ moment representations is impractical using these approaches.

\begin{table}[t]
\centering
  \caption{Per epoch training and inference time for Charades-STA dataset.}
  \label{tab:time}
  \begin{tabular}{l|c|c}
    \toprule
    Approach & Training time & Inference time\\
    \midrule
    Sliding-based & 35.05 s & 90.46 s\\
    HMAN & 21.18 s & 83.91 s \\
    \bottomrule
  \end{tabular}
  
\end{table}

TripSiam \cite{dong2018triplet} and DSLT \cite{lu2020deep} are two different variants of triplet loss which are used in object tracking. TripSiam defines a matching probability for each triplet to measure the possibility of assigning the positive instance to exemplar compared with the negative instance and tries to maximize the joint probability among all triplets during training. DSLT \cite{lu2020deep} utilizes modulating function to minimize the contribution of easy samples in the total loss. While both setups perform better than baseline approaches, we observe that there is a significant improvement in median retrieval rank (MR). This indicates that even if TripSiam and DSLT can not retrieve the correct moment, they are robust in terms of the semantic association between moments and sentences.

\underline{\textit{Effectiveness of Hierarchical Moment Encoder.}} HMAN utilizes stacked temporal convolutional layers in a hierarchical structure to represent video moments. We conduct experiments to analyze the effects of using the hierarchical moment encoder module in our proposed model. We consider two setups, i) \textbf{w/ TCN}: the hierarchical moment encoder module built using temporal convolutional network is present in the model and ii) \textbf{w/o TCN}: the hierarchical moment encoder module is replaced with a simple feedforward network to project the candidate moment representations in the joint embedding space. We consider both sum-margin based and max-margin based triplet loss to train the networks. Table \ref{tab:w_tcn} illustrates the effect of utilizing hierarchical moment encoder module. We observe that for both the learning approaches and for both datasets, there is a significant improvement in performance when the hierarchical moment encoder module is used. For example, in DiDeMo dataset, we observe $\sim14\%$ (sum-margin) and $\sim8\%$ (max-margin) absolute improvement in performance for $R@100, IoU=0.50$.

\underline{\textit{\update{Ablation Study of Learning Joint Embedding Space.}}} 
We conduct experiments to analyze the impact of different components of the loss function to learn the joint embedding space for our targeted task in DiDeMo dataset and reported the results in Table \ref{tab:embedding_didemo}. We use three setups to learn the joint embedding space:
\begin{itemize}[leftmargin=*]
    \item \textbf{HMAN (intra)}: Only uses $\mathcal{L}^{intra}$.  So the network only learns to distinguish intra-video moments.
    \item \textbf{HMAN (video)}: Only uses $\mathcal{L}^{video}$. So the network only learns to disntinguish moments from different videos based on global semantics.
    \item \textbf{HMAN (proposed)}: Our proposed approach, combination of $\mathcal{L}^{intra}$ and $\mathcal{L}^{video}$.  
\end{itemize}

In Table \ref{tab:embedding_didemo}, we observe that the performance of HMAN is poor for both the case of HMAN (intra) and HMAN (video). Performance of HMAN (intra) is better compared to HMAN (video) in Table \ref{tab:embedding_didemo} when higher $IoU$ threshold requirement is considered ($R@k, IoU=0.7$). This indicates that HMAN (intra) learns to better identify temporal boundaries in a video compared to HMAN (video), while HMAN (video) is better at distinguishing moments from different videos compared to HMAN (intra). However, when we combine both of these criteria, there is a significant performance boost as the model is able to effectively learn to identify both the correct video and the temporal boundary. All the results in Table \ref{tab:embedding_didemo} are reported for sum-margin based triplet loss setup.

\begin{figure}
    \centering
    \includegraphics[width =  .9\linewidth]{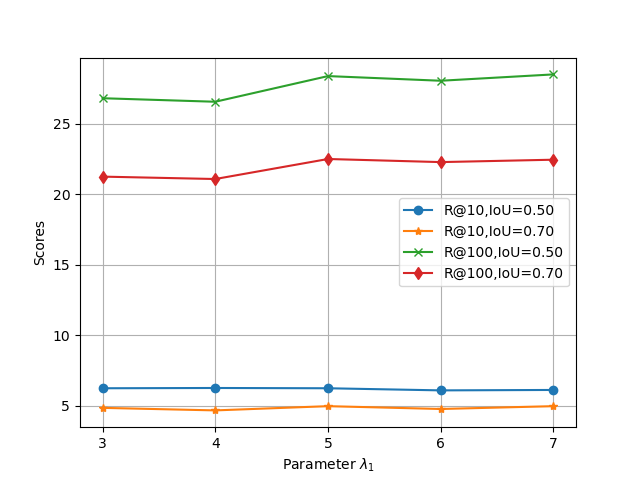}
    \caption{Illustration of $\lambda_1$ parameter sensitivity on the HMAN performance. We observe that for the set of values $\{3,4,5,6,7\}$, performance of HMAN is stable.}
    \label{fig:sensitivity}
\end{figure}

\underline{\textit{\update{Effectiveness of Utilizing Global Semantics.}}} \textcolor{black}{Our proposed learning objective utilizes global semantics to distinguish moments from different videos.}  
\textcolor{black}{To do so, we learn to align corresponding video-sentence pairs, where the video-sentence relevance $R(v,s)$ in the embedding space is computed in terms of moment-sentence similarity $S(m,s)$. So we use this video-sentence relevance score $R(v,s)$ to analyse the models performance to identify or retrieve the correct video given a text query and report the results in Table \ref{tab:R_retrieval}}. We use the standard evaluation criteria $R@k$ for video retrieval task and report $R@10$, $R@100$, and $R@200$ scores for DiDeMo and Charades-STA dataset. Here, $R@K$ calculates the percentage of query sentences for which the correct video is found in the top-K retrieved videos to the query sentence. In DiDeMo test set, there are $1{,}004$ videos with $4{,}016$ moment-sentence pairs ($\sim$ $4$ sentences per video) and in Charades-STA testset, there are $1{,}334$ videos with $3{,}720$ moment-sentence pairs ($\sim$ 2.8 sentences per video). Due to the one-to-many correspondences, we consider $4{,}016$ and $3{,}720$ video-sentence pairs respectively for DiDeMo and Charades-STA datasets for the video retrieval task, where a single video can pair up with multiple sentences. Table \ref{tab:R_retrieval} shows that both sum-margin (HMAN (sum)) and max-margin (HMAN (max)) based triplet loss setups of our proposed approach outperforms standard Visual Semantic Embedding based retrieval approach (VSE++) for the task of retrieving the correct video. Along with the consistent improvement of performance in all metrics for both datasets, We observe $\sim 40\%$ absolute improvement of retrieval performance for the metric R@200 for DiDeMo dataset. As the video-sentence relevance is computed in terms of moment-sentence similarity, this experiment validates the models capability to distinguish videos as well as moments from different videos utilizing global semantics.

\underline{\textit{\update{Analysis of Video Relevance Computation Approach.}}} In an untrimmed video with temporal language annotation, the segment/portion of the video mostly matches with the sentence semantics. So to compute the video-sentence relevance, it needs to focus on the moments that have higher similarity with the query sentence semantics. To tackle this issue, we compute the video-sentence relevance using LogSumExp pooling (Eqn. \ref{eqn:relevence}) of the moment-sentence similarity. In Table \ref{tab:R_type}, we analyze the significance of the LogSumExp pooling compared to average pooling for both sum-margin and max margin based triplet loss setups. In Table \ref{tab:R_type}, `ave' and `log' indicates average and LogSumExp pooling respectively, while `sum' and `max' indicates sum-margin based and max-margin based triplet loss respectively. For both DiDeMo and Charades-STA datasets, we observe that LogSumExp pooling performs better than average pooling for the target task of temporal localization of moments in video corpus in both sum-margin based and max-margin based triplet loss setups.

\underline{\textit{\update{Ablation Study of Different Visual Features.}}} We conduct experiments to study the performance of HMAN for different visual features for DiDeMo dataset and reported the results in Table \ref{tab:feature_table}. We use extracted features from VGGNet \cite{simonyan2014very}, ResNet-152 \cite{he2016deep} for RGB frames and optical flow features from \cite{wang2016temporal}.  In Table \ref{tab:feature_table}, we observe that a combination of RGB and optical flow features perform better than using only an RGB stream. It indicates the models increased capacity due to the increase in the number of learnable weights. As a result, HMAN is suitable to work with multiple encodings of the same data together compared to the shallow embedding networks \cite{anne2017localizing, escorcia2019temporal}. We have reported the results for sum-margin based triplet loss setup. 

\begin{figure}
    \centering
    \includegraphics[width =  \linewidth]{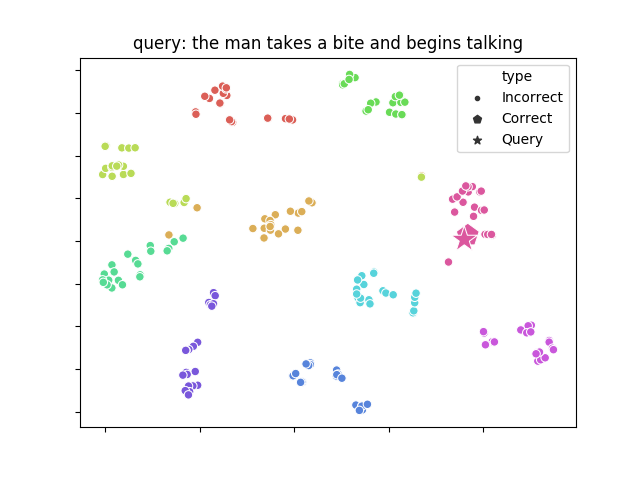}
    \caption{t-SNE visualization of text query representation and candidate moment representations. Different color represents different video. The color of the text representation is the same as the corresponding video. We use different markers for the representation of incorrect candidate moments, correct candidate moments and text. Here,
    representations of the text query and the correct candidate moment coincide. Also, the representations of candidate moments from the same video are clustered together.}
    \label{fig:tsne}
\end{figure}

\textit{\underline{Performance of HMAN on Decreased Number of Moment}-\underline{sentence Pairs.}} Since HMAN searches for the correct candidate moment across all the videos in the test set during inference, the temporal localization performance of HMAN is expected to improve by decreasing the number of moment-sentence pairs in the test set. We conduct experiments on DiDeMo dataset to evaluate the performance of HMAN (learned using sum-margin based triplet loss) on the decreased number of moment-sentence pairs in the test phase. We consider four setups: \textbf{HMAN (100\%)}: Model searches over the full test set during inference, \textbf{HMAN (50\%)}: Model searches over each 50\% of the test set separately and take the average of the scores, \textbf{HMAN (25\%)}: Model searches over each 25\% of test set separately and take the average of the scores, \textbf{HMAN (10\%)}: Model searches over each 10\% of test set separately and take the average of the scores. Table \ref{tab:decreasing} illustrates the performance for all four setups. We observe that with decreased number of test set moment-sentence pairs, the performance of HMAN improves.

\begin{figure*}
    \centering
    \includegraphics[width = \textwidth]{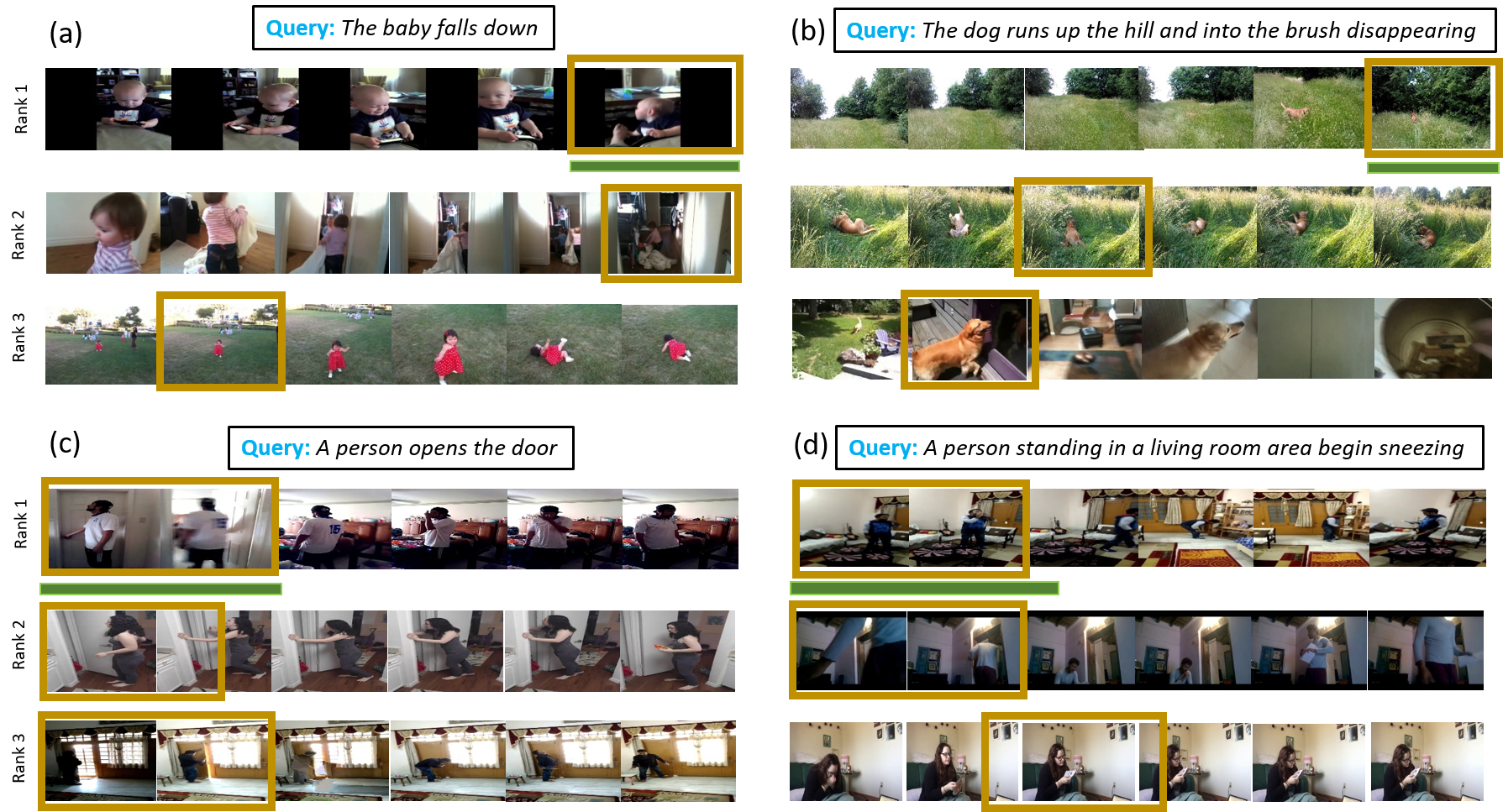}
    \caption{Example illustration of the performance of HMAN for the task of localization of moments in a corpus of videos. For each query sentence, we display the top-3 retrieved moments. The retrieved moments are surrounded by gold boxes and the ground truth moments are indicated by green lines. We observe that for each of the queries, the top-3 retrieved moments are semantically related to the sentence proving the efficacy of our approach. }
    \label{fig:qualitative}
\end{figure*}

\underline{\textit{Evaluation of Run Time Efficiency.}} We conduct experiments on the Charades-STA dataset to compare the run time of HMAN with the sliding window-based approaches. The differences in the sliding-based approach compared to the setup of HMAN is that: i) the moment encoder module with temporal convolutional network of HMAN is replaced by a simple single layer feedforward network, ii) instead of generating candidate moment representations directly from the video, we slide over the video to extract features of different temporal durations, then use extracted features to generate candidate moment representations. Table \ref{tab:time} illustrates that for both training case and inference case, the sliding-based approach takes longer than HMAN per epoch, even though the network is much smaller in the sliding-based approach compared to HMAN. For a fair comparison, we keep the number of candidate moments the same, and similar computations (apart from hierarchical moment encoder module replaced by single layer feed forward network) are done for both the approaches. We have computed the run time for five epochs and reported the average results. Here, the inference time is higher due to the added requirement of computing the cosine distance between each text query and all the candidate moment representations.

\underline{\textit{$\lambda_1$ Parameter Sensitivity Analysis.}} In our framework, $\lambda_1$ balances the contribution of $\mathcal{L}^{intra}$ and $\mathcal{L}^{video}$ for both sum-margin and max-margin case. We choose the value of $\lambda_1$ empirically. We conduct an experiment to check the sensitivity of HMAN performance based on a set of values for $\lambda_1$ in the DiDeMo dataset where $\lambda_1 \in \{3,4,5,6,7\}$. We observe that for this set of values of $\lambda_1$, the performance is stable.

\subsection{Qualitative Results}

\underline{\textit{t-SNE Visualization.}} We provide t-SNE visualization of embedding representations of text query and candidate moments in Fig. \ref{fig:tsne}. For a text query, we consider embedding representation of the text query, representations of candidate moments from the correct video, and representations of candidate moments from randomly picked nine other videos and visualize the distribution of representations. In Fig. \ref{fig:tsne}, different color represents different videos. Each video has $21$ candidate moments. We keep the color of the text query representation the same as the color of candidate moments representation from the correct video and use separate markers for correct candidate moment and text query representation. We 
observe that representations of the text query and the correct candidate moment coincide. Also, the representations of candidate moments from the same video are clustered together.

\underline{\textit{Example Illustration.}}
In Figure \ref{fig:qualitative}, we illustrate some qualitative results for our proposed approach. The two examples in the top row are for the DiDeMo dataset and the two examples in the bottom row are for the Charades-STA dataset. For each query sentence, we demonstrate the examples where the network is able to retrieve the correct moment as the rank-1 from the test set videos. We also display rank-2 and rank-3 moments retrieved by the model for each query sentence. Figure \ref{fig:qualitative}(a) shows that for the query \textit{`The baby falls down'}, the model was able to retrieve the correct moment with the highest matching. However, the interesting fact lies in the retrieved rank-2 and rank-3 moments. For the query \textit{`The baby falls down'}, the retrieved rank-2 and rank-3 moments also contain activity of a baby, including a baby falling down. Similar results are observed for other examples for both datasets. For example, in Figure \ref{fig:qualitative}(b), for the query sentence \textit{`A person opens the door'}, the model was able to retrieve the correct moment with the highest matching. However, all top-3 ranked moments contain activity related to a door. In the rank-2 moment, a person is opening a door and in the rank-3 moment, a person is fixing a door. Similarly, the top retrieved moments for a query of a dog running and hiding contain activities of a dog (Figure \ref{fig:qualitative}(b)) and top retrieved moments for a query of a person standing and sneezing contain standing activity and sneezing activity (Figure \ref{fig:qualitative}(d)). These results indicate the model's capability of retrieving moments with similar semantic concepts from the corpus of videos.

%% file: sections/N06_rv2_conclusion.tex
\section{Conclusion}
In this work, we explore an important and under-explored task of localizing moments in a video corpus based on text query. We adapt existing temporal localization of moments approaches and video retrieval approaches for the proposed task and identified the shortcomings of those approaches. Towards addressing the challenging task, we propose Hierarchical Moment Alignment Network (HMAN), a novel neural network that effectively learns a joint embedding space for video moments and sentences to retrieve the matching moment based on semantic closeness in the embedding space. Our proposed learning objective allows the model to identify subtle changes of intra-video moments as well as distinguish inter-video moments utilizing text-guided global semantic concepts of videos. We adopt both sum-margin based and max-margin based triplet loss setups separately and achieve performance improvement over other baseline approaches in both setups. We experimentally validate the effectiveness of our proposed approach on the DiDeMo, Charades-STA, and ActivityNet Captions datasets.